\documentclass[a4paper]{spie}

\usepackage[]{graphicx}
\usepackage{epsfig,portland}
\usepackage{afterpage}
\usepackage{fancyheadings,fancybox}
\usepackage{psboxit,pstricks,rotating}
\usepackage{psfrag}
\usepackage{ulem}
\usepackage{amsmath,bbm} 
\usepackage{amsfonts} 
\usepackage{amssymb}
\usepackage{floatflt}
\usepackage{theorem}
\usepackage{stmaryrd}
\usepackage{euscript}
\usepackage{pifont}
\usepackage{subfigure}
\usepackage{array}
\usepackage{epic,eepic}

\title{Restoration algorithms and system performance evaluation for active imagers}

\author{J\'er\^ome Gilles
\skiplinehalf
DGA/DET/CEP/GIP,\\ 16 bis, av Prieur de la C\^ote d'Or, 94114, ARCUEIL, France
}

\authorinfo{E-mail: jerome.gilles@etca.fr, Telephone: +33 (0)1 42 31 94 27}

\begin{document} 
\maketitle 

\begin{abstract}
This paper deals with two fields related to active imaging system. First, we begin to explore image processing algorithms to restore the artefacts like speckle, scintillation and image dancing caused by atmospheric turbulence. Next, we examine how to evaluate the performance of this kind of systems. To do this task, we propose a modified version of the german TRM3 metric which permits to get MTF-like measures. We use the database acquired during NATO-TG40 field trials to make our tests.
\end{abstract}


\section{INTRODUCTION}
\label{sec:intro}  
Since few years, active imaging systems appear as new imaging solutions. The fields of application are various: target recognition and identification, the search for people in non-cooperative zone, detection of optics,$\ldots$. The use of laser illumination permits to increase the identification range, the image resolution but some drawbacks also appear. Images are corrupted by different phenomenons like speckle, scintillation and image dancing (due to the laser propagation through the atmosphere).\\
In this paper, we propose to explore two different fields in relation with active imaging.\\
In the first part of this paper, we investigate the effect of atmosphere turbulence in the case of active imaging systems. The classical filter used in the litterature is the temporal mean filter. This filter gives good results but the edges of the objects are blurred. In order to improve the performance, we propose to use a temporal median filter. We show that, in a statistical point of view, the median filter is better adapted than the mean filter. Its main advantage is that it doesn't create new values which never appear during the acquisition process. We make different tests on the database acquired by the NATO-TG40 group. In the case of high turbulence, these filters are not efficient to correct the image dancing phenomenon. Then we present a new algorithm based on warping technics which we currently work on. The first results seems to be very promising.\\
In the second part of the paper, we propose a new method to evaluate the performance of this kind of systems. Our method use some ideas taken from the german TRM3 model. We define a new metric which uses directly the values measured at different locations in the image. This method permits to build a Modulation Transfert Function (MTF) like the one for passive systems. We conduct some experiments on the different systems used in the NATO-TG40 field trials and evaluate their performances.\\
We end the paper by giving some conclusions and perspectives for the future.

\section{Active imaging systems}
\begin{figure}
\begin{center}
\includegraphics[scale=0.15]{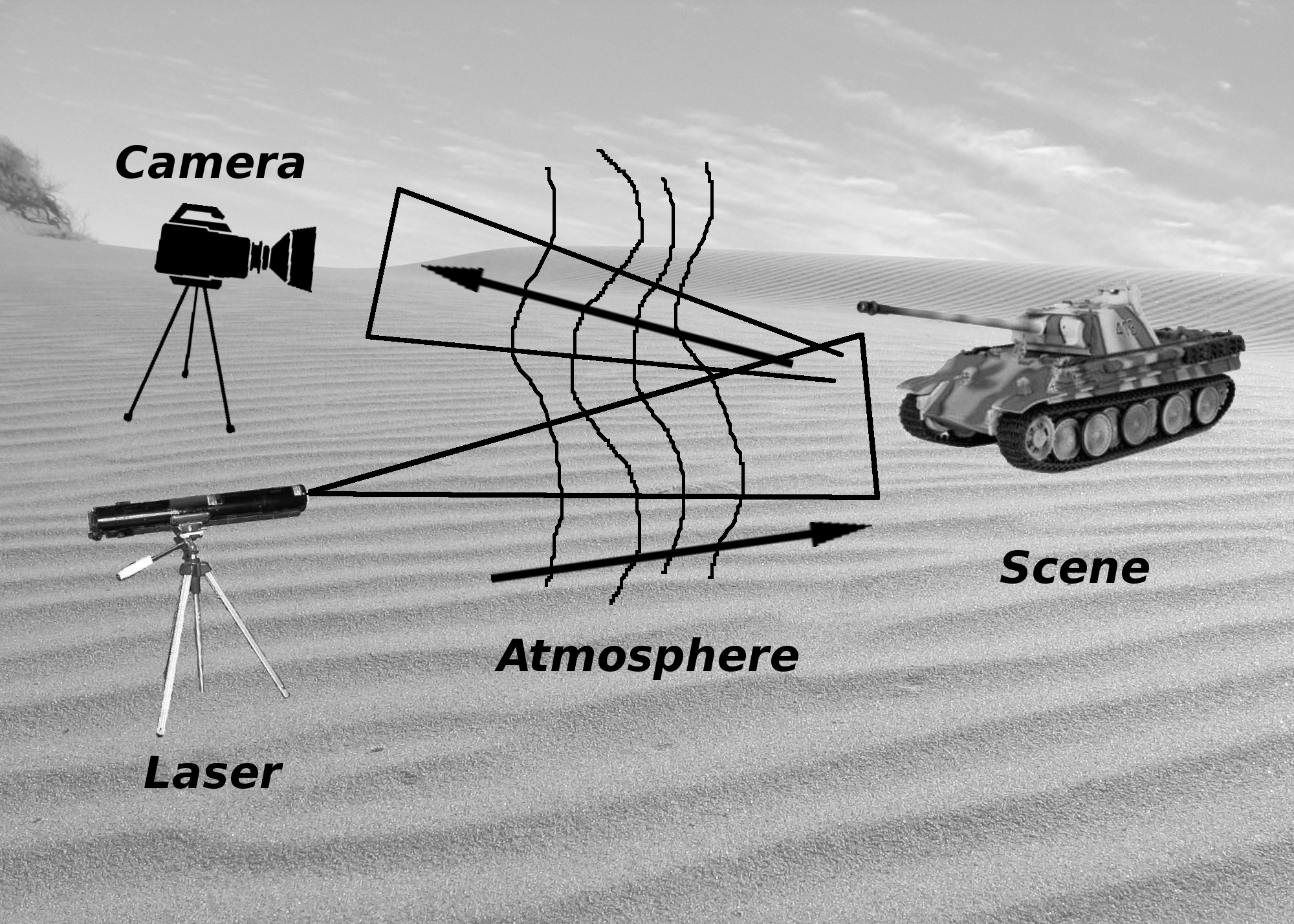}
\end{center}
\caption{Active imaging principle.}
\label{fig:imact}
\end{figure}

The principle of active imaging consists to acquire the reflected laser which illuminate the scene (see figure \ref{fig:imact}). The advantage is that this technique permits to increase the range capability of imaging systems. But the laser is submited to the influence of the atmosphere (back-propagation, diffusion process, turbulence, $\ldots$), to the reflexion effect on the different objects present in the scene and particularly the scintillation effect. These drawbacks need to be treated by adequate processing.

\section{Restauration algorithms}
\label{sec:algo}
In this section we first examine the mean and temporal median filter. Next we describe some current work which use the diffeomorphism mathematical concept to do some warping on the images.

\subsection{Temporal median filter}
First lets define some notations. We write $\{I_n\}_{0 \leqslant n \leqslant N}$ an image sequence, where $N$ is the number of frames in the sequence. We assume that images are size of $I\times J$ and the coordinate of a pixel is written $(i,j)$ (then the image domain is $\Omega=\llbracket 0;I \rrbracket\times \llbracket 0;J \rrbracket$).\\
In the litterature, people usually use a simple temporal mean filter:
\begin{equation}
\forall (i,j)\in\Omega \qquad I_{mean}(i,j)=\frac{1}{N}\sum_{n=0}^N I_n(i,j)
\end{equation}
This filter works well but it blured the image, specially at the high frequencies (near the edges in the images). In order to increase the performance we propose to use the temporal median filter:

\begin{equation}
I_{med}(i,j)= MED((I_n(i,j))_{0\leqslant n\leqslant N}) \quad \text{we assume that $N$ is odd}
\end{equation} 
We recall that this filter consists of rearranging the vector
\begin{equation}
\left(I_0(i,j), I_1(i,j), \ldots, I_{N-1}(i,j)  \right)
\end{equation}
in an increasing way, then we get the vector
\begin{equation}
\left(\tilde{I}_0(i,j), \tilde{I}_1(i,j), \ldots, \tilde{I}_{N-1}(i,j)  \right)
\end{equation}
where $\tilde{I}_n(i,j)\leqslant\tilde{I}_{n+1}(i,j)$. As $N$ is assume to be odd, the median value is $\tilde{I}_{\frac{N-1}{2}}(i,j)$.\\

The advantage of the median filter is that it doesn't create new values, the median value is a value which really appears many time during the acquisition process. This behaviour is interesting because it produces less blur than the average filter along edges. Let's illustrate this property by the following example. Assume we have a signal which can have only two values: $a$ and $b$. Assume these values appear $N_a$ and $N_b$ times respectively during the acquisition process. Then the outputs of both filters are given by:
\begin{equation}
I_{mean}=\frac{1}{N_a+N_b}(aN_a+bN_b)
\end{equation}
and
\begin{equation}
I_{med}=
\begin{cases}
a \quad \text{if}\quad N_a>N_b \\
b \quad \text{else}
\end{cases}
\end{equation}

\begin{figure}
\includegraphics[width=\textwidth]{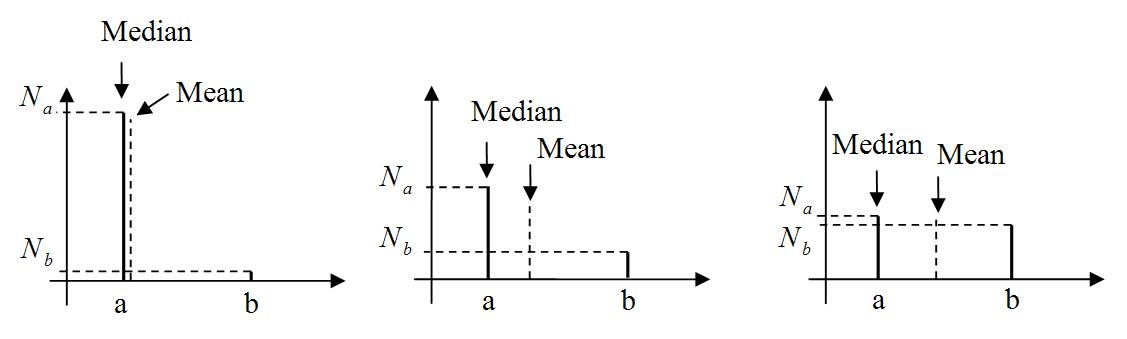}
\caption{Behaviour of the average and median filters.}
\label{fig:behaviour}
\end{figure}

This example is illustrated in figure \ref{fig:behaviour}. More higher 
the turbulence is, more $N_b$ increases and $N_a$ decreases. Then we can clearly see that the behaviour of the median filter is better adapted than the mean filter. At each time it gives the right value unlike the average filter which gives a compromise.

We mention that the mean and median filters can be declined in a time-shifted version. We mean that we use only $P\leqslant N$ frames and then we shift in time the filter step by step (in order to get a final sequence of the same size of the original one, we assume the Dirichlet condition on time).\\

\begin{figure}[!h]
\begin{center}
\includegraphics[scale=0.23]{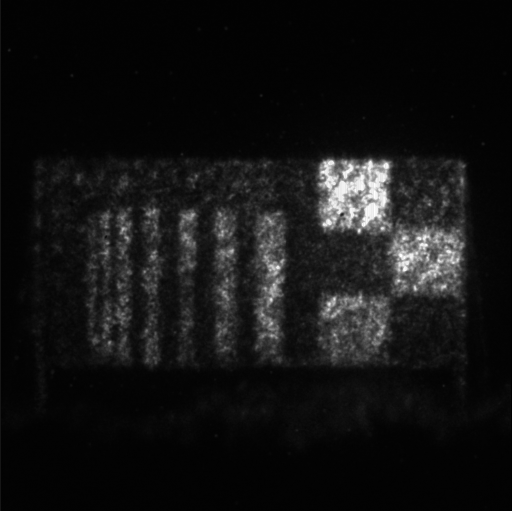}
\includegraphics[scale=0.23]{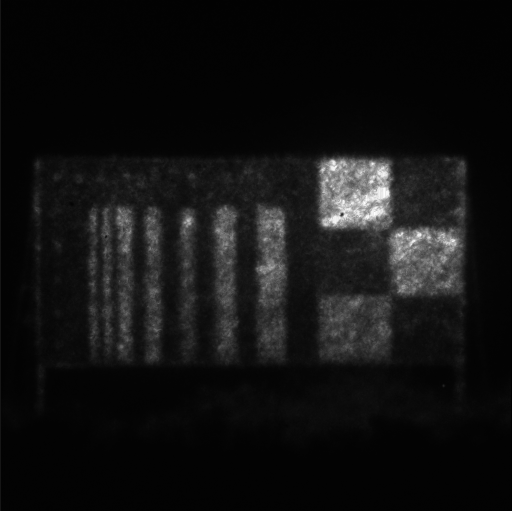} 
\includegraphics[scale=0.23]{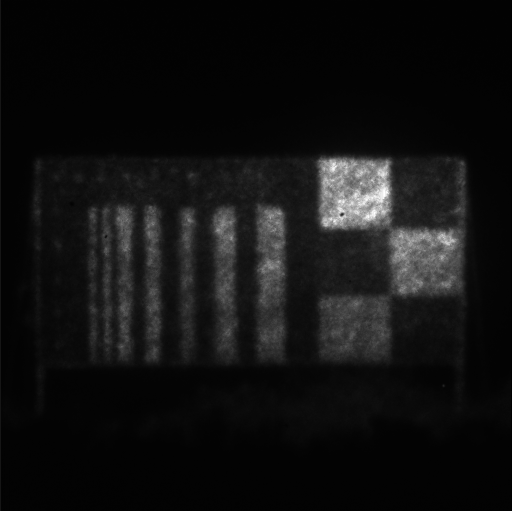} 
\includegraphics[scale=0.23]{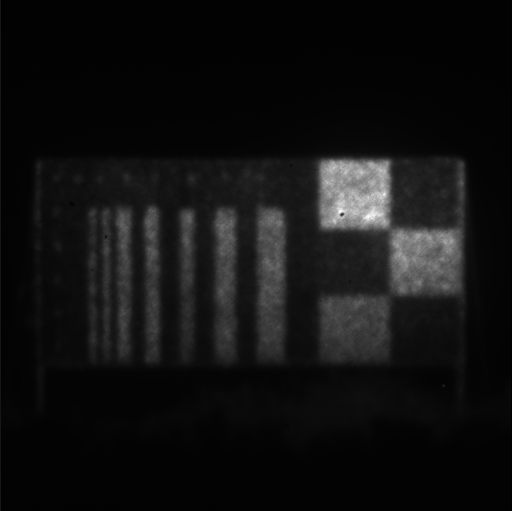} 
\end{center}
\caption{Temporal median filter. From left to right: one image extracted from the original sequence, results obtained with $N=5$, $N=10$ and $N=50$ respectively.}
\label{fig:meduk1}
\end{figure}

\begin{figure}[!h]
\begin{center}
\includegraphics[scale=0.23]{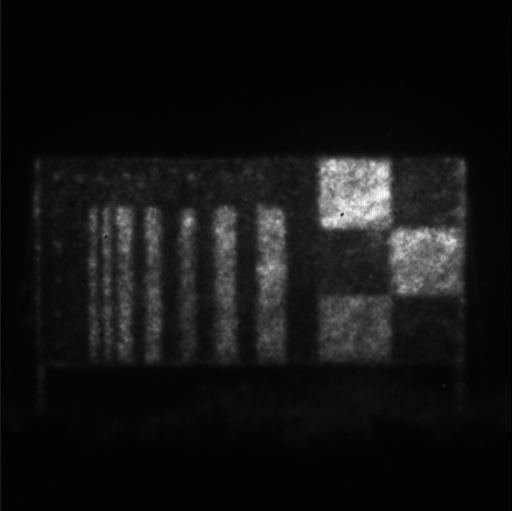}
\includegraphics[scale=0.23]{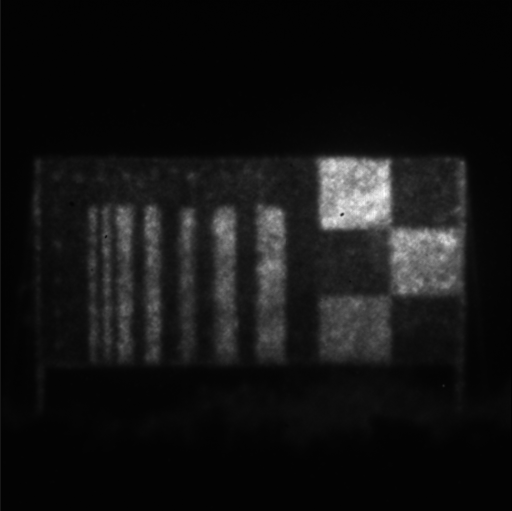}
\includegraphics[scale=0.23]{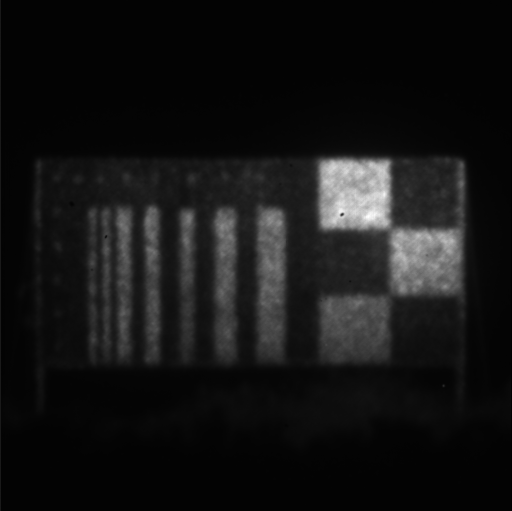}
\end{center}
\caption{Temporal mean filter. From left to right: results obtained with $N=5$, $N=10$ and $N=50$ respectively.}
\label{fig:moyuk1}
\end{figure}

We made our experiments on the database acquired during the field trials made by the NATO-TG40 group. This group works on the modelization of active imaging systems. These trials were made at the White Sands Missile Range (US, New Mexico). The figure \ref{fig:meduk1} shows the results we get with the temporal median filter for different values of $N$ applied on a sequence acquired by the UK system. We clearly see that the results are very interesting even with only $N=5$. See figure \ref{fig:moyuk1} to compare with the results of the mean filter. We can see that, as we expected, the high frequencies are better restored by the median filter than the mean filter. In general, during our experiments with different level of turbulence, we saw that the median filter works much well than the mean filter.\\

Just mention here the work of Lemaitre et al.\cite{mlemaitre,mlemaitre2} where the authors use some local filtering technics like adaptive Wiener filter or Laplacian filter. This approach gives good results but the complexity of the algorithms is harder and the processing time is higher than the mean or median filters.\\

Even if the results are good, we can see that in the case of high turbulence none of these methods corrects the geometric deformation (image dancing) induced by the atmosphere movement. The next section presents some work we currently do in order to deal with this problem.

\subsection{Geometric deformation correction}
\begin{figure}[!h]
\begin{center}
\includegraphics[scale=0.7]{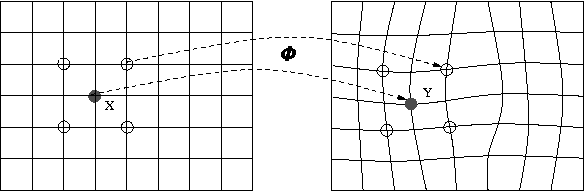}
\end{center}
\caption{Warping principle}
\label{fig:warping}
\end{figure}

In this subsection, we present some work we currently do on the correction of image dancing. The image dancing is difficult to modelize because the phenomenon is clearly random. The approach we choose is to use an elastic image deformation technique based on diffeomorphism. We recall that a function $\phi:M \rightarrow N$ is a diffeomorphism if both $\phi$ and $\phi^{-1}$ are differentiable. \\
We just describe here the principle of the method we use for image dancing correction, more details will be published in a futur paper. Suppose we search for a diffeomorphism $\phi_k$ which permits to warp an image $I_k$ to a reference image $I_R$ (we mean that $I_R=I_k\circ \phi$, see figure \ref{fig:warping}). 

Beg\cite{beg} shows that it exists a velocity field defined by equation (\ref{eq:energie}).

\begin{equation}\label{eq:energie}
\hat{v}=\arg\min_{v\in L^2([0,T],V)} E(v)=\frac{1}{2}\int_0^T \Vert v_t\Vert_V^2dt + \frac{C}{2}\Vert I_0 \circ \phi_{T,0}^v -I_R\Vert_{L^2}^2
\end{equation} 

Then the diffeomorphism is related to $v$ by the following relation:
\begin{equation}
\phi_{0,t}^{v}=\phi_{0,0}^{v}+\int_0^t v_s\circ \phi_s ds.
\end{equation}

All details on diffeomorphic elastic methods could be found in the work of Beg\cite{beg,beg2005}, Miller\cite{miller}, Younes\cite{younes2000,younes2001}.\\

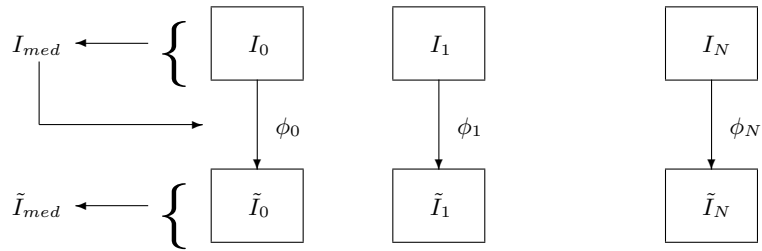
\begin{figure}[!h]
\begin{center}
\setlength{\unitlength}{3947sp}%
\begingroup\makeatletter\ifx\SetFigFont\undefined%
\gdef\SetFigFont#1#2#3#4#5{%
  \reset@font\fontsize{#1}{#2pt}%
  \fontfamily{#3}\fontseries{#4}\fontshape{#5}%
  \selectfont}%
\fi\endgroup%
\begin{picture}(4658,1692)(1936,-1776)
{\color[rgb]{0,0,0}\thinlines
\put(5226,-399){\circle*{70}}
}%
{\color[rgb]{0,0,0}\put(5396,-399){\circle*{70}}
}%
{\color[rgb]{0,0,0}\put(5565,-399){\circle*{70}}
}%
{\color[rgb]{0,0,0}\put(5735,-399){\circle*{70}}
}%
{\color[rgb]{0,0,0}\put(5226,-1416){\circle*{70}}
}%
{\color[rgb]{0,0,0}\put(5396,-1416){\circle*{70}}
}%
{\color[rgb]{0,0,0}\put(5565,-1416){\circle*{70}}
}%
{\color[rgb]{0,0,0}\put(5735,-1416){\circle*{70}}
}%
{\color[rgb]{0,0,0}\put(5226,-908){\circle*{70}}
}%
{\color[rgb]{0,0,0}\put(5396,-908){\circle*{70}}
}%
{\color[rgb]{0,0,0}\put(5565,-908){\circle*{70}}
}%
{\color[rgb]{0,0,0}\put(5735,-908){\circle*{70}}
}%
{\color[rgb]{0,0,0}\put(3193,-626){\framebox(565,452){}}
}%
{\color[rgb]{0,0,0}\put(4322,-626){\framebox(565,452){}}
}%
{\color[rgb]{0,0,0}\put(6017,-626){\framebox(565,452){}}
}%
{\color[rgb]{0,0,0}\put(2798,-399){\vector(-1, 0){452}}
}%
{\color[rgb]{0,0,0}\put(3476,-626){\vector( 0,-1){564}}
}%
{\color[rgb]{0,0,0}\put(4605,-626){\vector( 0,-1){564}}
}%
{\color[rgb]{0,0,0}\put(6300,-626){\vector( 0,-1){564}}
}%
{\color[rgb]{0,0,0}\put(3193,-1642){\framebox(565,452){}}
}%
{\color[rgb]{0,0,0}\put(4322,-1642){\framebox(565,452){}}
}%
{\color[rgb]{0,0,0}\put(6017,-1642){\framebox(565,452){}}
}%
{\color[rgb]{0,0,0}\put(2120,-513){\line( 0,-1){395}}
\put(2120,-908){\vector( 1, 0){1017}}
}%
\put(3589,-964){\makebox(0,0)[lb]{\smash{{\SetFigFont{9}{10.8}{\rmdefault}{\mddefault}{\updefault}{\color[rgb]{0,0,0}$\phi_0$}%
}}}}
\put(4718,-964){\makebox(0,0)[lb]{\smash{{\SetFigFont{9}{10.8}{\rmdefault}{\mddefault}{\updefault}{\color[rgb]{0,0,0}$\phi_1$}%
}}}}
\put(6412,-964){\makebox(0,0)[lb]{\smash{{\SetFigFont{9}{10.8}{\rmdefault}{\mddefault}{\updefault}{\color[rgb]{0,0,0}$\phi_N$}%
}}}}
\put(3420,-1472){\makebox(0,0)[lb]{\smash{{\SetFigFont{9}{10.8}{\rmdefault}{\mddefault}{\updefault}{\color[rgb]{0,0,0}$\tilde{I}_0$}%
}}}}
\put(4549,-1472){\makebox(0,0)[lb]{\smash{{\SetFigFont{9}{10.8}{\rmdefault}{\mddefault}{\updefault}{\color[rgb]{0,0,0}$\tilde{I}_1$}%
}}}}
\put(6242,-1472){\makebox(0,0)[lb]{\smash{{\SetFigFont{9}{10.8}{\rmdefault}{\mddefault}{\updefault}{\color[rgb]{0,0,0}$\tilde{I}_N$}%
}}}}
\put(6242,-456){\makebox(0,0)[lb]{\smash{{\SetFigFont{9}{10.8}{\rmdefault}{\mddefault}{\updefault}{\color[rgb]{0,0,0}$I_N$}%
}}}}
\put(4549,-456){\makebox(0,0)[lb]{\smash{{\SetFigFont{9}{10.8}{\rmdefault}{\mddefault}{\updefault}{\color[rgb]{0,0,0}$I_1$}%
}}}}
\put(3420,-456){\makebox(0,0)[lb]{\smash{{\SetFigFont{9}{10.8}{\rmdefault}{\mddefault}{\updefault}{\color[rgb]{0,0,0}$I_0$}%
}}}}
\put(1951,-1472){\makebox(0,0)[lb]{\smash{{\SetFigFont{9}{10.8}{\rmdefault}{\mddefault}{\updefault}{\color[rgb]{0,0,0}$\tilde{I}_{med}$}%
}}}}
\put(1951,-456){\makebox(0,0)[lb]{\smash{{\SetFigFont{9}{10.8}{\rmdefault}{\mddefault}{\updefault}{\color[rgb]{0,0,0}$I_{med}$}%
}}}}
\put(2855,-569){\makebox(0,0)[lb]{\smash{{\SetFigFont{41}{49.2}{\rmdefault}{\mddefault}{\updefault}{\color[rgb]{0,0,0}\{}%
}}}}
\put(2855,-1585){\makebox(0,0)[lb]{\smash{{\SetFigFont{41}{49.2}{\rmdefault}{\mddefault}{\updefault}{\color[rgb]{0,0,0}\{}%
}}}}
{\color[rgb]{0,0,0}\put(2798,-1416){\vector(-1, 0){452}}
}%
\end{picture}%
\end{center}
\caption{Diffeomorphic correction algorithm}
\label{fig:diffalgo}
\end{figure}

In our case, we want to correct the geometric deformations induced by the turbulence phenomenon by diffeomorphic warping. The goal is to deform all the image sequence to a reference image. The first question is: how can we get a reference image? We simply choose the result given by the median filter ($I_R=I_{med}$). Then we warp all the sequence to $I_{med}$ (see figure \ref{fig:diffalgo}). We get the new restorated image $\tilde{I}_{med}$ by applying a second time the temporal median filter to the warped sequence $\{ \tilde{I}\}_{0 \leqslant n \leqslant N}$. In theory, this process could be iterated many times but in practice one step seems to be sufficient. \\

\begin{figure}[!h]
\begin{center}
\begin{tabular}{ccc}
\includegraphics[scale=2]{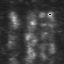} &
\includegraphics[scale=2]{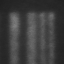} &
\includegraphics[scale=2]{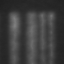} \\
Original image $I_k$ & $I_{med}$ & $\tilde{I}_{med}$ 
\end{tabular}
\end{center}
\caption{Diffeomorphic correction algorithm results.}
\label{fig:diffeo}
\end{figure}

In figure \ref{fig:diffeo}, we see the improvement given by this approach. The first image is one extracted from the original sequence, the second is the result we get by applying the temporal median filter as in the previous subsection. The last one is the result obtained when applying the median filter to the warped sequence. As we expected, the geometry is more regularized and the image dancing effect is diminished.

But some blur remains in $\tilde{I}_{med}$. We currently explore different ways to deblur the images. The final step could be a resolution improvement by the use of techniques like super-resolution methods.

\section{System performance evaluation}
\begin{figure}[!h]
\begin{center}
\includegraphics[scale=0.6]{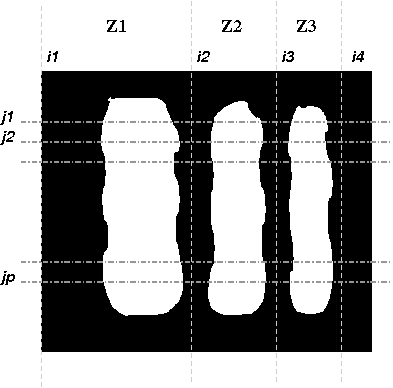}
\end{center}
\caption{Image partition principle.}
\label{fig:zonemire}
\end{figure}

In this part of the paper we investigate the problem of system performance measurement. There are two aspects of performance measurement, the first is the prediction of the performance. In that case, the goal is to validate the technology choices and the parameters used to build a system in regard with desired performances. To do this task, we need to modelize the optronic chain and particularly the modelization of the atmosphere turbulence but it is a difficult task and lots of work are currently done around the world.\\
The second aspect is how to evaluate the performances of an existing system? In that case, we assume we only have the images acquired by the system. The goal is to determine the ``quality'' of the image, both before and after all potential processings.\\
Our work is based on the Average Modulation at Optimum Phase (AMOP) idea used in the german TRM3 model\cite{amop}. The goal of this method is to estimate a pseudo MTF directly from the measured images representing a barchart. In AMOP, one of the main assumption is that the measure can be done globally or locally on the image to take care about some nonlinear effects such as aliasing. For the images which are deformed by the turbulence effect, we propose to enforce the locality of the measure by taking it at different lines in the image (this allows to evaluate local distortions). Figure \ref{fig:zonemire} shows how we partitionned the image in order to make the measures. We will call this method NonLinear Amplitude Measure (NLAM).

Now, let's define some notations: we extract $p$ lines, the current line will be noted $j$. We also define different zones, $Z_k$, which correspond to a spatial frequency (one cycle). Then $s_j^k$ will be the signal extracted from $Z_k$ at line $j$. The minimum and the maximum of $s_j^k$ are denoted respectively $\overline{S_j^k}$ and $\underline{S_j^k}$. Then we adapt the average gain definition used in the AMOP model for each frequency by

\begin{equation}
\Delta S_k=\frac{1}{p}\sum_{j=1}^p \left|\overline{S_j^k}-\underline{S_j^k}\right|
\end{equation}

\begin{figure} 
\begin{center}
\includegraphics[width=0.45\textwidth]{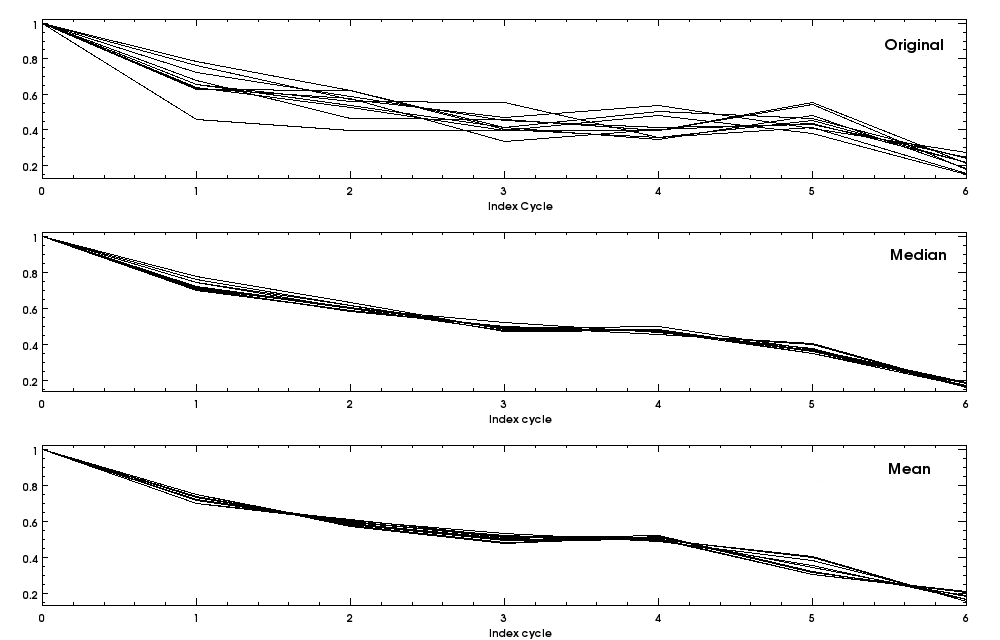}
\includegraphics[width=0.45\textwidth]{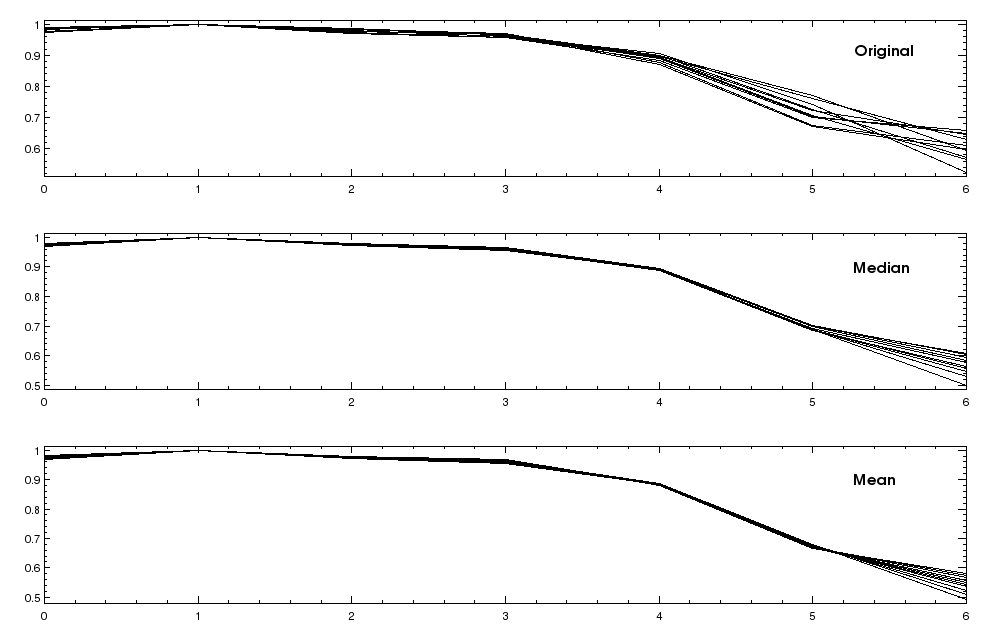}
\end{center}
\caption{NLAM measures obtained on the ten first frames of the swedish (left) and canadian (right) sequences.}
\label{fig:nlamET}
\end{figure}

\begin{figure} 
\begin{center}
\includegraphics[width=0.45\textwidth]{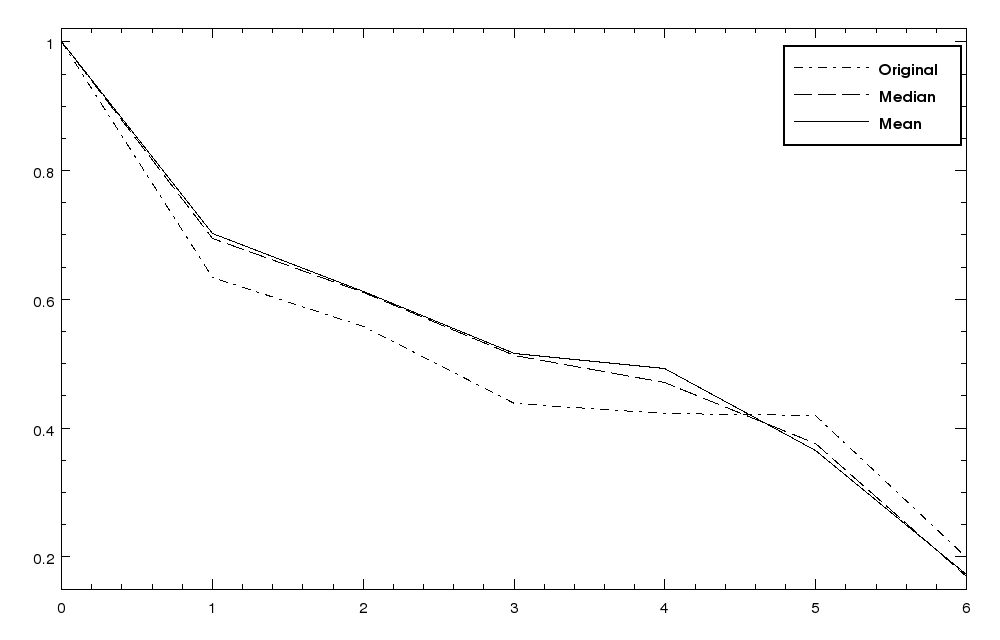}
\includegraphics[width=0.45\textwidth]{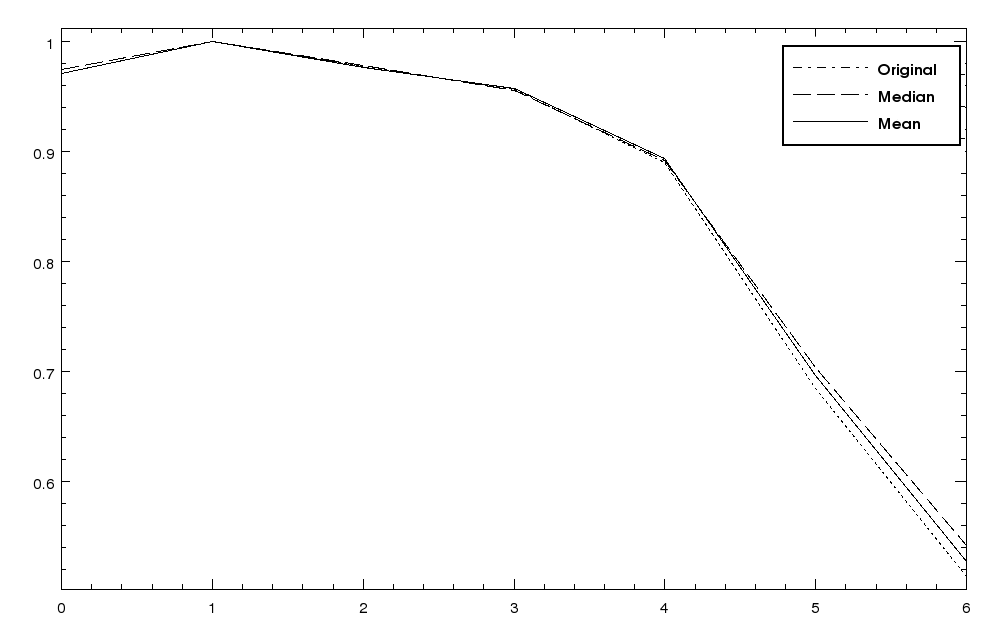}
\end{center}
\caption{Average NLAM measures obtained on the swedish (left) and canadian (right) sequences.}
\label{fig:nlamMOY}
\end{figure}

Then we build the graph $(k,\overline{\Delta}S_k)$ as in the AMOP model, this corresponds to NLAM. This graph behaves like a MTF and permits to evaluate the performances of the system. \\
We test this method on two kinds of sequence: one acquired by the swedish active imaging system and the second acquired by a Canadian high speed passive camera. Figure \ref{fig:nlamET} gives the measures we get on the ten first frames of each original sequence, the mean filtered one and the median filtered one. Figure \ref{fig:nlamMOY} gives the mean of all graphs, we consider it as the pseudo MTF of the system. As we expected, the NLAM has the behaviour of an MTF. We clearly see that the median filter gives better results in the high frequencies than the mean filter.

\section{Conclusion}
In this paper, we adress two problems related to active imaging systems: image restoration and system performance evaluation. First we show that the temporal median filter is better than the temporal mean filter. But none of these filters take care about the image dancing phenomenon. Then we proposed to use a warping technic based on diffeomorphism. This method permits to improve the regularity of the geometry in the image.\\
Next we adress the problem of system performance evaluation. We propose a metric (named NLAM) based on a modified version of the german AMOP metric. NLAM gives us a MTF like function.\\
Future work will be dedicated to validation of the NLAM method on a bigger part of the NATO TG40 dataset. We also want to compare the measured performance with performance predicted by simulation models. Other future work will concern the restoration algorithms which need to be evaluated. To do this, we will take an original non degraded image, add the effect of the atmosphere on it by simulation (for example by the model described by Potevin et al\cite{potevin}) and then apply the restoration processing. At the end, we can compare the original and the restored images.

\section*{Acknowledgements}
We would like to thank the other members of the NATO TG40 group for their interest and the very interesting discussions we had. We also would like to thank Tristan Dagobert (DGA/DET/CEP/GIP) who implemented all the diffeomorphic algorithms and made the tests.

\bibliographystyle{spiebib}

\bibliography{spie2007}

\end{document}